\documentclass[review]{elsarticle}

\usepackage[noend]{algpseudocode}
\usepackage{algorithmicx,algorithm}
\usepackage{xcolor}
\usepackage{amsmath,amssymb,amsfonts}
\usepackage{graphicx}
\usepackage{hyperref}

\journal{Journal of \LaTeX\ Templates}

\begin{document}

\begin{frontmatter}

\title{Synthetic Hard Negative Samples for Contrastive Learning}

\author{Hengkui Dong}

\author{Xianzhong Long\corref{mycorrespondingauthor}}
\cortext[mycorrespondingauthor]{Corresponding author}
\ead{lxz@njupt.edu.cn}

\author{Yun Li}
\author{Lei Chen}
\address{School of Computer Science $\&$ Technology, School of Software \\ Nanjing University of Posts and Telecommunications, Nanjing, China, 210023}

\begin{abstract}
Contrastive learning has emerged as an essential approach for self-supervised learning in visual representation learning. The central objective of contrastive learning is to maximize the similarities between two augmented versions of an image (positive pairs), while minimizing the similarities between different images (negative pairs). Recent studies have demonstrated that harder negative samples, i.e., those that are more difficult to differentiate from the anchor sample, perform a more crucial function in contrastive learning. This paper proposes a novel feature-level method, namely sampling synthetic hard negative samples for contrastive learning (SSCL), to exploit harder negative samples more effectively. Specifically, 1) we generate more and harder negative samples by mixing negative samples, and then sample them by controlling the contrast of anchor sample with the other negative samples; 2) considering the possibility of false negative samples, we further debias the negative samples. Our proposed method improves the classification performance on different image datasets and can be readily integrated into existing methods.
\end{abstract}


\end{frontmatter}


\section{Introduction}

During the recent years,  self-supervised learning has undergone notable advancements, with the emergence of numerous noteworthy methodologies. Among these, contrastive learning has garnered significant attention, owing to its exceptional performance in many downstream tasks.

\begin{figure}[ht]
\centering
\includegraphics[width=1.0\textwidth]{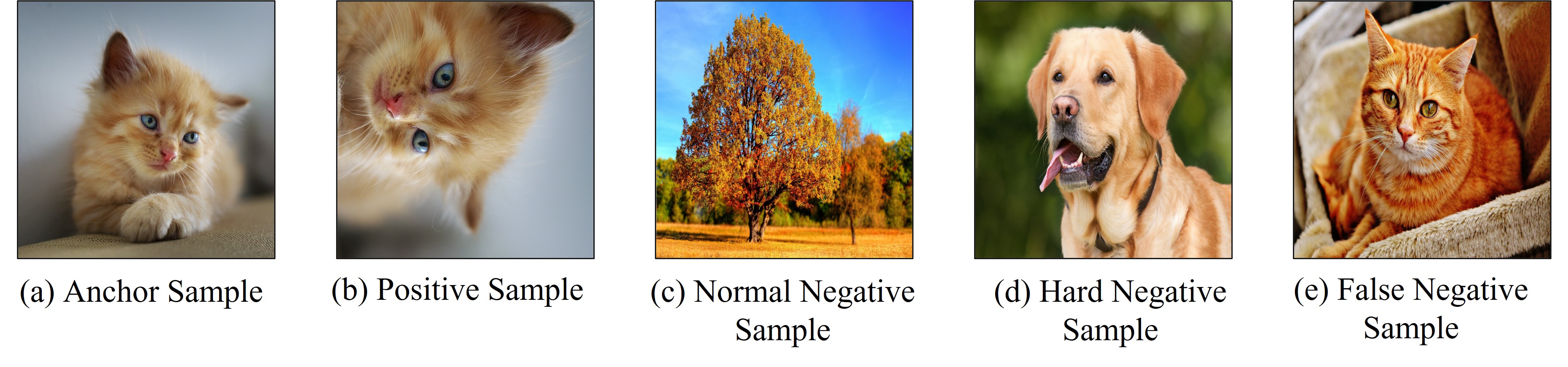}
\vspace{-3 em}
\caption{Illustrations of (a) Anchor Sample, (b) Positive Sample, (c) Normal Negative Sample, (d) Hard Negative Sample and (e) False Negative Sample.}
\label{fig1}
\end{figure}

Contrastive learning is mainly utilized to train the network model by maximizing the similarities of positive pairs and minimizing the similarities of negative pairs. Among many of the existing methods, positive samples usually come from different augmentation versions of the same image, as shown in Figure \ref{fig1}(b). In contrast, negative samples are obtained either from the current batch or memory bank. Because the number of negative samples will directly affect the training results, in view of this, SimCLR provides more negative samples by increasing the batch size \cite{b1}, while MoCo \cite{b2} and MoCo v2 \cite{b3} using momentum update mechanism keeps a queue as memory bank, which can dynamically maintain the count of negative samples.

However, more negative samples do not necessarily mean harder negative samples. The study \cite{b4} found that only a small minority of negative samples are both necessary and sufficient for contrastive learning. The queue in momentum contrastive learning contains many normal negative samples and few hard ones (about 5$\%$). Only few hard negative samples actually work when the model is trained, while many existing methods have ignored the importance of hard negative samples. In Figure \ref{fig1}, it is observed that the normal negative sample and the anchor sample belong to different classes, and their semantic information is dissimilar. Furthermore, the hard negative sample has similar semantic information with the anchor sample, although they belong to different classes.

In this study, we emphasize the significance of hard negative samples in the training process of contrastive learning.
We propose to sample synthetic hard negative samples. Firstly, instead of generating negative samples by increasing the batch size or memory bank, we select the hardest existing negative samples in the feature space and synthesize new ones. Secondly, we control the weights of negative samples in loss function by applying weighting factors to negative samples. Finally, 
we also utilize the debiased sampling to deal with the issue of false negative sample in contrastive learning. The false negative sample in Figure \ref{fig1}(e) contains the similar semantic information to the anchor sample but is considered as a negative sample.

All our contributions are listed as follow:
\begin{itemize}
\item A method for sampling synthetic hard negative samples in contrastive learning is proposed. It enables the network model to learn more challenging representations from harder negative samples, by generating more and harder negative samples and sampling synthetic hard negative samples in the feature space.

\item  
Considering the issue of the false negative sample, we also debias negative samples to ensure the truthfulness of the negative samples.

\item The proposed approach is compatible with existing contrastive learning methods with only few computational overheads, and is achieved by simple modifying the standard contrastive loss function.

\end{itemize}

\section{Relation work}
\subsection{Contrastive learning}
Contrastive learning as a typical method of discriminative self-supervised learning has made  incredible progress in the field of visual representation learning. Both positive  and negative samples are essential and directly affect the performance of the model in contrastive learning.

For an input image, typically using a certain data augmentation to generate positive pairs. SimCLR applies transformations to control the images from the perspective of their raw pixel values, e.g., random cropping, color jittering, etc, which can modulate the hardness of self-supervised task \cite{b1}.
Moreover, NNCLR \cite{b18} samples nearest-neighbour samples from the feature space as additional positive samples, which yields more information than data augmentation. MYOW \cite{b19} also believes that images from data augmentation are not sufficiently diverse, therefore samples that are similar to but different from the anchor sample are sought as positive samples.

For negative pairs, the selection of negative samples has received less attention in contrastive learning. In general, the model performance is improved by increasing the count of negative samples.  MoCo \cite{b2,b3} maintains a queue of negative samples as memory bank,  which can provide numerous negative samples for the network model.

Additionally, there are also some methods that are unlike those above. SwAV \cite{b7} combines contrastive learning with clustering, rather than comparing with a huge amount of negative samples, while they utilize prior information and compare with clustering centers. BYOL \cite{b5} does not use any negative samples and applies a prediction head to prevent the model from collapsing. In addition, SimSiam \cite{b6} is a further simplification of BYOL, and can achieve more effective results through stop-gradient update.

\subsection{Negative Sample Sampling}
MoChi \cite{b8} selects negative samples based on the similarity with anchor sample and mixes them with anchor and negative samples. Likewise, a negative sample sampling strategy is formed to mine hard negative pairs by normalized $L_{2}$  distance with anchor sample \cite{b9}.

What's more, to reduce the effect of false negative sample, \cite{b10} presents a debiased contrastive loss and analyses the true negative samples distribution. Furthermore, \cite{b11} proposes a negative samples sampling method that assigns weighting factor to each negative pair according to similarity. Meanwhile, in order to seek better negative samples and debias false negative samples into true negative samples, the method proposed in \cite{b10} was also applied in \cite{b11}. \cite{b20} uses the importance sampling for correcting the bias of random negative samples and designs the sampling distribution under the Bayesian framework. In \cite{b12}, negative samples in the memory bank are treated as learnable weights and attack against the network model by viewing them as adversarial samples. Therefore, the network model is able to better capture the information of negative samples and eventually learn a more robust representation.

However, the existing methods have some problems, simple random sampling does not necessarily provide useful semantic information. The selected and generated negative samples may be false negative samples when selection strategy based on similarity is used to generate negative samples. If only sampling negative samples, the number of hard negative samples may not be sufficient. The proposed SSCL method not only generates more and harder negative samples but also imposes weighted sampling on them again after ensuring the truthfulness of negative samples.

\section{Approach}
Contrastive learning attempts to pre-train a network model by comparing the similarities and dissimilarities between image feature vectors. Consequently, similar samples of the same category are likely to be near one another in feature space, while dissimilar samples from different categories are expected to be more distant from each other.

Specifically, given an input image $x$, it is turned into a pair of augmented images $t\left( x \right)$ and ${t}'\left( x \right)$, which can be considered  as a positive pair. A backbone encoder $f_{\theta}$ and a projector head $g_{\theta}$ are used to map the positive pair into feature vectors, 
$z = g_{\theta} \left( f_{\theta} \left( t\left( x \right) \right) \right)$ and 
${z}' = g_{\theta} \left( f_{\theta} \left( {t}'\left( x \right) \right) \right)$.
Then, the comtrastive loss function can be defined in the following form:

\begin{figure*}[ht]
\centering

\includegraphics[width=1.0\textwidth]{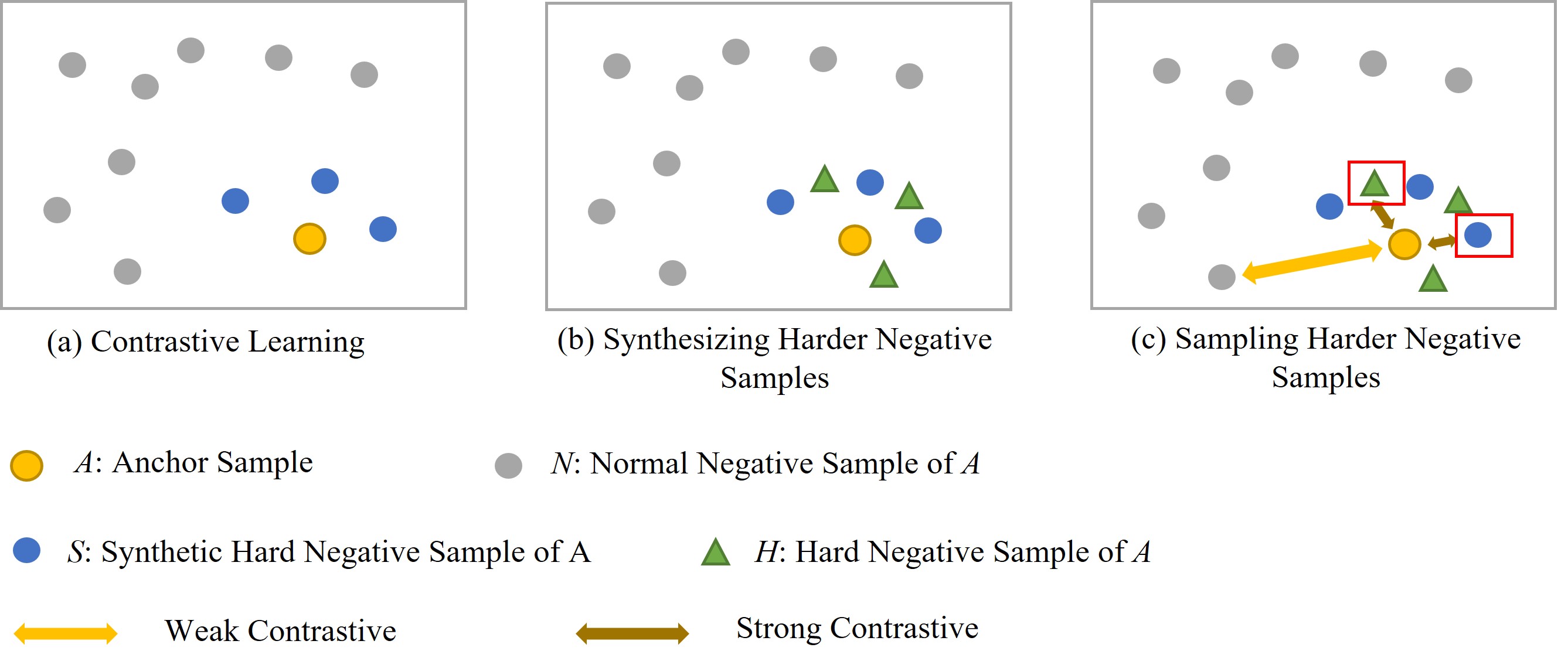}
\vspace{-3 em}
\caption{Illustrations of (a) Contrastive Learning; (b) Synthesizing Harder Negative samples; (c) Sampling Harder Negative samples. In (a), the normal negative samples ($N$) are far from anchor sample ($A$) in the feature space, and the hard negative samples ($H$)  which contain more semantic information about ($A$)  are closer to ($A$). In (b), we take advantage of ($H$) to generate synthetic hard negative samples ($S$) by linear combination. In (c), the anchor sample ($A$) has stronger negative contrast with harder negative samples around it, and we select harder negative samples (red box) to provide more semantic information to the network model.}
\label{fig2}
\end{figure*}

\begin{equation}
\begin{aligned}
\mathcal{L} _{con}= 
 -\log\frac{  \exp \left(  sim\left ( z,{z}’      \right  )/r       \right )  }    {\exp \left(  sim\left ( z,{z}’     \right  )/r       \right )
     +\sum\limits_{{z}’'\epsilon \left\{ Z_{neg} \right\}}^{}\exp\left(sim\left ( z,{z}’'      \right  )/r       \right )}     \label{eq}
\end{aligned}
\end{equation}
where $z$ serves as the feature vector of the anchor sample, ${z}’$ and ${z}’'$ are the feature vectors of the anchor's positive pair and negative pairs respectively. $\small\{ Z_{neg} \small\}$ denotes the set of negative samples, which contains feature vectors of different negative samples. The cosine similarity is usually adopted as the similarity function $sim$, and $r$ is the temperature parameter for the loss function. The distribution of negative samples on feature space in contrastive learning is shown in Figure \ref{fig2}. In Figure \ref{fig2}(a), most negative samples are far from the anchor sample, with only a few hard negative samples around the anchor sample.

\begin{figure*}[ht]
\centering

\includegraphics[scale=0.43]{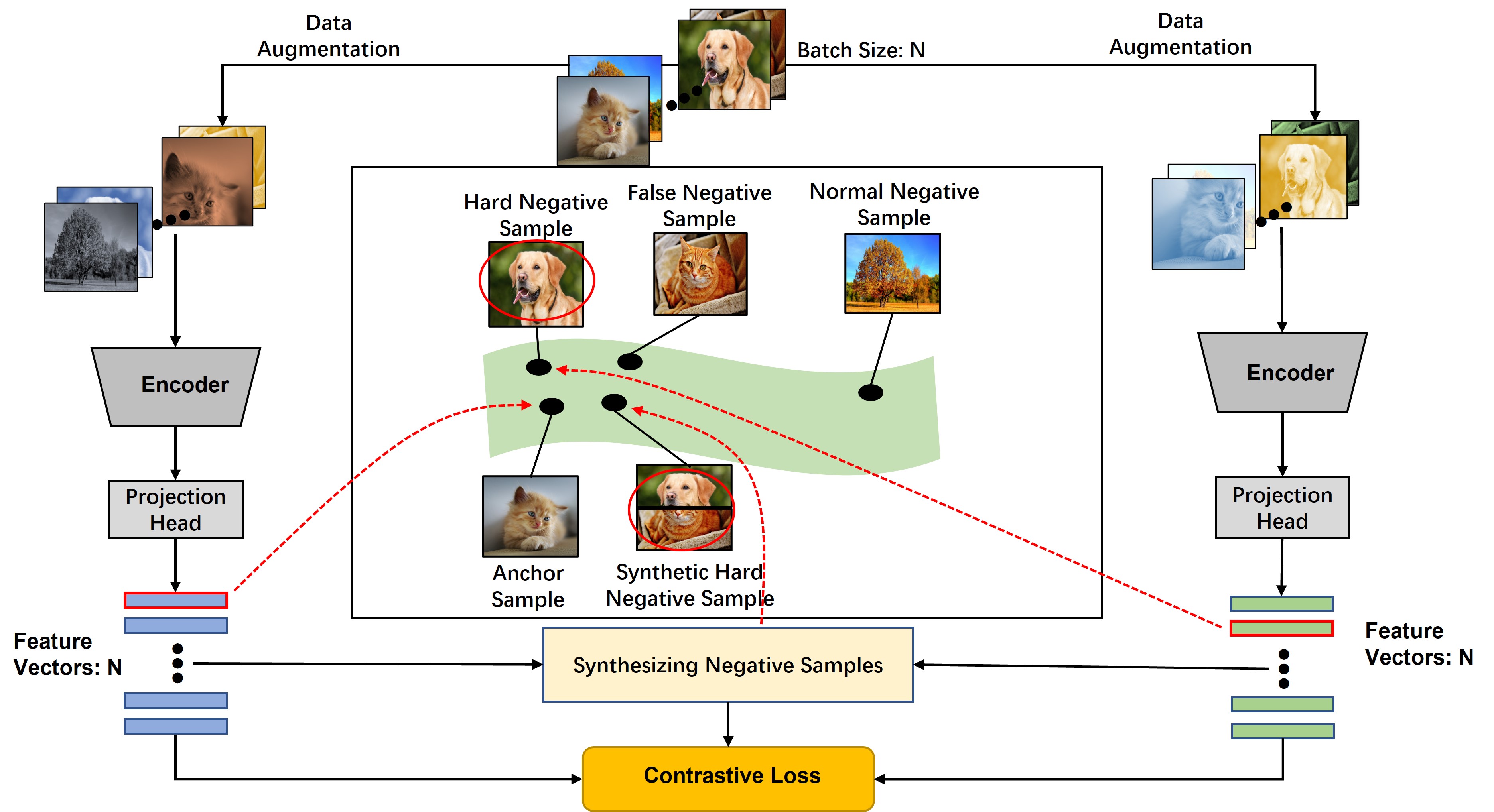}
\caption{The architecture of SSCL. Given a batch of N images, we first apply data augmentation and then encode and projection head to obtain two batch feature vectors (2N). For one image, we generate new harder negative samples by synthesizing negative samples from the 2N-2 remaining images. The red dashed arrow indicates the projection of the feature vector. We then combine the original samples with the newly synthesized ones to compute the contrastive loss. }
\label{fig3}
\end{figure*}

\subsection{Generating More Synthetic Harder Negative Samples}

For negative sample set  $\left\{ Z_{neg} \right\}=\left\{{z_{1}}''\cdots,{z_{n}}'' \right\}$, we adopt a strategy of selecting the hardest negative sample features from $\left\{ Z_{neg} \right\}$ based on their similarities to the anchor sample feature ${z}$. Specifically, we define the hardest negative sample set $\small\{\widetilde{Z}_{neg} \small\}=\small\{\widetilde{{z_{1}}}''\cdots,\widetilde{{z_{s}}}'' \small\}$, where $\small\{\widetilde{Z}_{neg} \small\}\subseteq \left\{ Z_{neg} \right\}$, and
$sim\small ( z,\widetilde{{z_{i}}}''      \small  ) >  sim\small ( z,\widetilde{{z_{j}}}''      \small ), {\forall}i<  j$.

Next, to further enhance the diversity of negative samples, we generate more synthetic harder negative samples by linear combination of the hardest negative samples, let $\left\{ H_{neg} \right\}=\left\{h_{1}\cdots,h_{k} \right\}$ as the set of synthetic hard negative sample, $h_{p}$ would be given by:
\begin{equation}
h_{p}=\alpha_{p} \widetilde{{z_{i}}}'' +\left(  1-\alpha_{p}  \right) \widetilde{{z_{j}}}''
\label{equ2}
\end{equation}where $\alpha_{p} \epsilon \left(0,1 \right)$, $\widetilde{{z_{i}}}''$ and $\widetilde{{z_{j}}}''$are chosen from $\small\{\widetilde{Z}_{neg} \small\}$  randomly.
$\left\{ H_{neg} \right\}$ as an expansion of the $\left\{ Z_{neg} \right\}$ provides more synthetic harder negative samples for the network model. The count of negative samples in $\small\{\widetilde{Z}_{neg}\small\}$ is $s$ and in $\left\{ H_{neg} \right\}$ is $k$. In Figure \ref{fig2}(b), the proposed method for synthesizing harder negative samples supplies more and harder negative samples around anchor samples.

\subsection{Sampling  Harder Negative Samples }

Besides generating more synthetic hard negative samples, we also assign a weighting factor to each negative pair. We adopt each pair’s similarity as its weight, i.e., $w_{z_i,z_j} = \exp \left(  sim\left ( z_i,z_j      \right  )/r       \right )  $, in this way, negative samples with greater similarities to anchor sample have higher weights. Meanwhile, We additionally define weighting control factor $\beta$ of $w_{z_i,z_j}$ to control the hardness of the weights, namely, $\widetilde{w}_{z_i,z_j}=\beta \cdot w_{z_i,z_j} $.
By incorporating these weights, the proportion of hard negative samples will become greater in the negative sample set,
harder negative samples play a more significant role in the loss function. In this way, the network model can pay closer attention to and learn more essential representations.

However, the negative sample set commonly used and the newly created synthetic negative sample set are taken from the entire data distribution $p(x)$, thus there may be false negative samples in the set, i.e., there will be samples in the set of negative samples that may constitute positive pair with the anchor sample. To avoid such a problem, we adopt the idea of debiased contrastive learning \cite{b10,b11} from positive unlabeled-learning \cite{b13} to sample hard negative samples without access to true labels.

Positive unlabeled-learning \cite{b13} decomposes the entire data distribution as: $p(x)= \tau p^{+}(x)+(1-\tau)p^{-}(x) $, where $p^{+}(x)$ and $p^{-}(x)$ respectively indicate the sample distribution from the same and different label of anchor sample $x$, $\tau$ is the class probability which denotes that samples from $p(x)$ have the same label as $x$. Then, $p^{-}(x)$ can be written as: $p^{-}(x)=\small(p(x)-\tau p^{+}(x)\small)/(1-\tau)$, we approximately estimate  $p^{-}(x)$ by  taking advantage of $p(x)$ and $\tau$.

Finally, we rearrange the negative sample term in contrastive loss function (\ref{eq}), i.e., rewriting $\sum\limits_{{z}''\epsilon \left\{ Z_{neg} \right\}}^{}\exp\left(sim\small ( z,{z}''      \small  )/r       \right )$ as a new term:
\begin{equation}
\frac{1}{1-\tau} \Big( 
\sum\limits_{{z}''\epsilon \left \{ Z_{neg}  \right \} \cup  \left \{ H_{neg}  \right \}   }^{}  \widetilde{w}_{z_i,z_j}  \cdot \exp \left(  sim\small ( z,{z}''    \small  )/r       \right )    
\\ - \tau\cdot  M\cdot  \exp \left(  sim\small ( z,{z}'      \small  )/r       \right )        \Big)
\label{equ3}
\end{equation}
where $\widetilde{w}_{z_i,z_j}$ is the weights for negative samples, $M$ is the count  of negative samples, and unlike the previous contrastive learning, our approach makes feature of negative sample  ${z}''$ come from $\left \{ Z_{neg}  \right \} \cup  \left \{ H_{neg}  \right \} $. In Figure \ref{fig2}(c), allowing the model to learn stronger representations by sampling harder negative samples that have stronger contrast with anchor sample. Eventually, we can obtain the complete loss function:



\begin{equation}
\begin{aligned}
\label{equ4}
\resizebox{1.0\hsize}{!}{$\mathcal{L}_{SSCL}$=
 $-\log\frac{  \exp \left(  sim\left ( z,{z}’      \right  )/r       \right )  }    {\exp \left(  sim\left ( z,{z}’     \right  )/r       \right )
    +\frac{1}{1-\tau} 
    \left( \sum \limits_{{z}''\epsilon \left \{ Z_{neg}  \right \} \cup  \left \{ H_{neg}  \right \}}   \widetilde{w}_{z_i,z_j}  \cdot \exp \left(  sim\small ( z,{z}''    \small  )/r       \right )    - \tau \cdot  M \cdot  \exp \left(  sim\left ( z,{z}'  \right  )/r       \right )        \right) }$}
\end{aligned}
\end{equation}

Our proposed SSCL method is implemented on the framework of SimCLR. The architecture of our approach is shown in Figure \ref{fig3} and is similar to SimCLR. We get the negative samples from the current batch and further synthesize negative samples utilizing feature vectors. All sample features are eventually used for contrastive learning.

Additional details of SSCL are presented in Algorithm \ref{al} .

\begin{algorithm}[]
    \caption{: Pseudocode of SSCL, PyTorch-like}
    \label{al}
     {\bf Input:}
    
     \quad  $f_{\theta }$ and $g_{\theta }$: backbone encoder and projection head  
    
    \quad   $r$: temperature parameter   \quad     E: total epoch number 

     \quad     k: the number of negative samples in $\left\{H_{neg} \right\}$ 
     
    \quad  $\beta$: weighting control factor, $\tau$: class probability

\begin{algorithmic}[1]
\For{epoch=0 to E} 
   \For{{\bf all} $ x$ in batch }   \qquad \qquad \qquad      \quad //sample a batch of N images
   
\State $z_{1}$=$g_{\theta }( f_{\theta }(t(x)))$   \qquad \qquad \qquad      \qquad //compute projections
  
    \State   $z_{2}$=$g_{\theta }( f_{\theta }({t}'(x)))$  \qquad \qquad \qquad      \qquad// projection: N$\times$C
    \State //compute exp of similarity for positive samples
    \State pos = exp(sum($z_1*z_2$,dim=-1)/$r$)// pos:N$\times$1
    
    \State pos = cat([pos,pos])  \qquad    \qquad \qquad //pos:2N$\times$1
    \State //compute similarity matrix 
    \State out = cat($[z_1$,$z_2$])   \qquad    \qquad \qquad \quad    // out:2N$\times$C
    
    \State n = exp((out,out.T())/$r$) \qquad  \qquad // n:2N$\times$2N \\
    
      \State Synthesize new negative sample set $\left \{ Z_{neg}  \right \} \cup  \left \{ H_{neg}  \right \} $ using Eq.(\ref{equ2})
      
      \State neg = Synthesize(n)\qquad  \qquad \qquad// neg:2N$\times$(2N-2+k)\\
      
       \State Sample negative samples using Eq.(\ref{equ3})
       
       \State $\widetilde{w} = \beta $*neg \qquad  \qquad \qquad  \qquad \qquad //compute weighting factor
       
     \State Neg=($\widetilde{w}$*neg-(2N-2+k)*$\tau$*pos).sum()/(1-$\tau$) // Neg:2N$\times$1

    \EndFor

    \State$\mathcal{L}_{SSCL}$=
 $-\log\frac{ \rm pos  }    {\rm pos
    +\rm Neg}$
 
    \State  update networks $f_{\theta }$ and $g_{\theta }$ to minimize $\mathcal{L}_{con}$
\EndFor

\State \Return backbone encoder $f_{\theta }$, and throw away $g_{\theta }$ 

\end{algorithmic}
\end{algorithm}

\section{Experiments}
\subsection{Experimental settings}
\subsubsection{Datasets}

In order to demonstrate the validity of our proposed approach, we present experimental results across different datasets and training strategies of model. Most training models for contrastive learning are performed on large-scale datasets like ImageNet \cite{b16}. 
The experiments of our proposed method are mainly evaluated on several different datasets, including CIFAR10 \cite{b14}, CIFAR100 \cite{b14} and TinyImageNet \cite{b15}.

We employ the training set for pre-training task and the test set for the downstream linear evaluation in CIFAR10 and CIFAR100.
The linear evaluation results are used as evaluation criteria for the methods. For TinyImageNet, we randomly selected 10 000 images from the training datase as the validation set, and the remaining 90 000 images are treated as the training set. Evaluation of the performance on the validation set is ultimately regarded as the experimental results. Table \ref{table1} provides a concise overview of each of the three datasets.


\begin{table}[h]
\centering
\caption{Description of the benchmark datasets}
 \vspace{+1.0em}
\label{table1}
\resizebox{0.95\columnwidth}{!}{
\begin{tabular}{ccccc}
\hline
\textbf{Dataset}      & \textbf{Training set} & \textbf{Test rest} & \textbf{Image size } & \textbf{Number of categories} \\ \hline
CIFAR10      & 50 000       & 10 000                  & 32$\times$32   & 10   \\
CIFAR100     & 50 000       & 10 000                   & 32$\times$32 & 100   \\
TinyImageNet  & 100 000      & 10 000                   & 64$\times$64   & 200    \\ \hline

\end{tabular}}
\end{table}

\begin{table}[h]
\centering
\caption{Experimental parameter settings.}
 \vspace{+1.0em}
 \label{table5}
\begin{tabular}{cccc}
\hline
                      & \textbf{CIFAR10} & \textbf{CIFAR100} & \textbf{TinyImageNet} \\\hline

Batch size            & 256     & 256      & 256          \\
Epoch                 & 200     & 200      & 200          \\

Warmup epoch          & 20      & 20       & 20           \\
Learning rate         & 0.1     & 0.1      & 0.1          \\
Warmup learning rate          & 10    &  10      &  10          \\
Temperature parameter & 0.5     & 0.5      & 0.5          \\
($\beta$, $\tau$)              & (1.0, 0.1) & (1.0, 0.05) & (1.0, 0.05)     \\
(s, k)                   & (32, 8)    & (32, 8)     & (32, 8)        \\ \hline
\end{tabular}
\end{table}

All experiments in this study are performed on an Nvidia GTX 3090 GPU.

\subsubsection{Default Setting}
 We use ResNet-18 \cite{b17} as the encoder architecture with 512-dimension feature, and employ projection head into the lower 128-dimension feature in all experiments. The models in all experiments are trained for 200 epochs, with batch size of 256, and temperature parameter of 0.5.
 \paragraph{Pre-training}

We employ stochastic gradient descent (SGD) algorithm with learning rate of 0.1 (learning rate = 0.1$\times$Batch Size$/$256) and weight decay of $10^{-3}$.  Furthermore, we utilize linear warmup for the first 20 epochs with learning rate of $10^{-4}$ and decrease the learning rate with cosine decay schedule. 
The selection of the optimal parameters for pre-training is partially shown in Table \ref{table5}.

For MoCo-related experiments, the momentum coefficient $m$ and the queue size are set to 0.99 and 16385 respectively.

 In our proposed method, to generate synthetic hard negative samples, we set the number of negative samples in $\small\{\widetilde{Z}_{neg} \small\}$ and $\left\{H_{neg} \right\}$ to 32 and 8 (s = Batch Size$/$8, k = Batch Size$/$32). According to the original literature \cite{b11}, we set weighting control factor $\beta$ to 1.0, and for CIFAR10, we set class probability $\tau$ to 0.1. For CIFAR100 and TinyImageNet, since they have more classes, we set smaller $\tau$ to 0.05.

\paragraph{Evaluation}
Following \cite{b1,b2,b8,b11}, we adopt linear fine-tuning strategies on freezing encoder and exchange the projection head with a linear layer to evaluate the effectiveness of pre-training. The linear classifier is trained for 100 epochs within a cosine decay approach using the  SGD algorithm, with learning rate of 10, weight decay of 0, momentum of 0.9, and batch size equal to pre-training.

\begin{table}[h]
\caption{Comparison under the linear evaluation criteria on three datasets }

\label{table2}
\begin{center}
\resizebox{1.0\columnwidth}{!}{
\begin{tabular}{llcclcllll}
\hline
\multicolumn{1}{c}{\textbf{Method}} &  & \multicolumn{2}{c}{\textbf{CIFAR10}}       &  & \multicolumn{2}{c}{\textbf{CIFAR100}} & \multicolumn{1}{c}{} & \multicolumn{2}{l}{\textbf{TinyImageNet}}  \\ \cline{3-4} \cline{6-7} \cline{9-10} 
\multicolumn{1}{c}{}       &  & top-1 & \multicolumn{1}{l}{top-5} &  & top-1         & top-5        & \multicolumn{1}{c}{} & top-1 & \multicolumn{1}{c}{top-5} \\ \hline

SimCLR\cite{b1}                     &  & 82.35   & 99.32                       && 52.02         & 82.12        &                      & 25.22 & 52.20                     \\
MoCo v2\cite{b3}                    &  & 81.45   & 99.19                      & & 47.71         & 78.64        &                      & 22.57 & 48.87                     \\
BYOL\cite{b5}                       &  & 86.15   &  99.51                     & & 55.36         & 84.23        &                      & 13.34 & 34.07                     \\
MoCHi\cite{b8}                      &  & 82.49   &    99.26                    & & 51.49         & 81.36        &                      & 26.93 & 53.94                     \\
DCL\cite{b10}                         &  & 84.56   &   99.38                  &  & 53.75         & 83.03        &                      & 25.56 & 52.83                     \\
HCL\cite{b11}                         &  & 85.91   &  \textbf{99.53}                      &  & 58.71         & 85.46        &                      & 37.15 & 63.94                     \\
SimCLR-SSCL                       &  & \textbf{86.20}   &  99.37                    &  & \textbf{60.01}         & \textbf{86.34}        &                      & \textbf{37.31} & \textbf{64.15}                     \\ \hline
\end{tabular}}
\end{center}
\end{table}

\subsection{Comparison with related methods}
Table \ref{table2} shows the linear classification results of our method SimCLR-SSCL and other relevant methods.
We evaluate the accuracy of an image classification utilizing top-1 and top-5 accuracy. 
Specifically, the top-1 accuracy is the ratio of instances where the model's prediction matches the true label exactly. The top-5 accuracy is the ratio of instances where the true label is ranked among the top 5 predicted categories by the model.

Based on Table \ref{table2}, it can be seen that the accuracy of SSCL on three datasets almost exceeds that of other methods. The top-1 accuracy of SSCL on the CIFAR100 dataset is 1.3 percentage points higher than the existing best method HCL. This shows the effectiveness of our method, SSCL  facilitates the model to focus more on harder negative samples that contain useful semantic information, thereby contributing to the enhancement of the network model's performance.


\begin{table}[ht]
\caption{Ablation studies for SSCL based on SimCLR}
\vspace{+1.0em}
\centering
\label{table3}
\resizebox{0.95\columnwidth}{!}{
\begin{tabular}{ccllcllcl}
\hline
\textbf{Method}      & \multicolumn{2}{c}{\textbf{CIFAR10}} &  & \multicolumn{2}{c}{\textbf{CIFAR100}} &  & \multicolumn{2}{c}{\textbf{TinyImageNet}} \\ \cline{2-3} \cline{5-6} \cline{8-9} 
            & top-1        & top-5        &  & top-1         & top-5        &  & top-1           & top-5          \\ \hline
SimCLR      & 82.35        & 99.32        &  & 52.02         & 82.12        &  & 25.22           & 52.20          \\
w/synthesis & 82.55        & 99.44        &  & 51.88         & 81.33        &  & 25.28           & 52.33          \\
w/synthesis+debias & 84.78        & 99.39        &  & 53.24         & 82.65        &  & 26.16           & 53.39          \\
w/sampling(HCL \cite{b11})    & 85.91        & \textbf{99.53}        &  & 58.71         & 85.46        &  & 37.15           & 63.94          \\
SimCLR-SSCL & \textbf{86.20}        & 99.37        &  & \textbf{60.01}         & \textbf{86.34}        &  & \textbf{37.31}          & \textbf{64.15}          \\ \hline
\end{tabular}}
\end{table}

\begin{figure}
\centering
\includegraphics[scale=0.1]{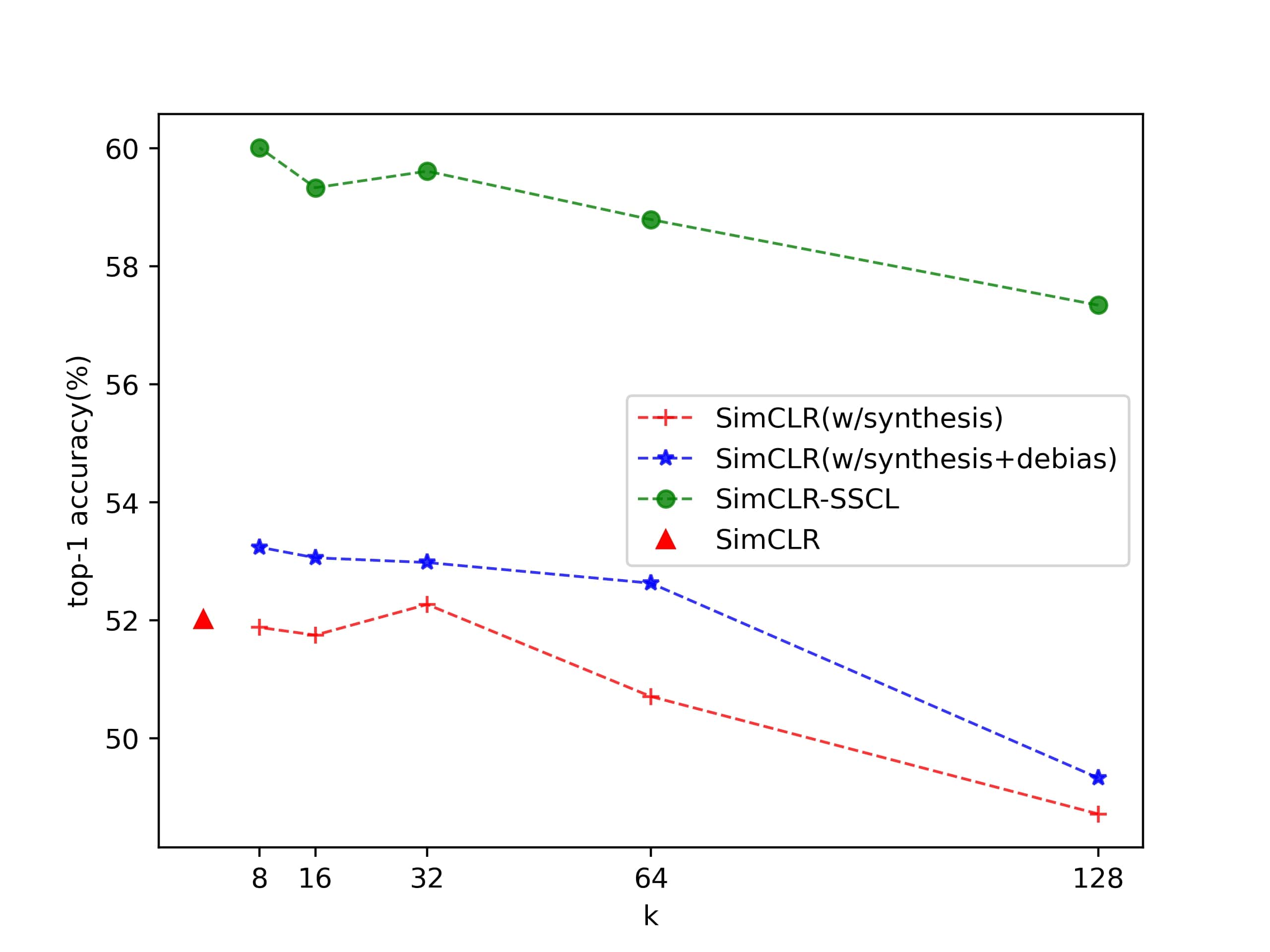}
 \vspace{-1em}               
\caption{Top-1 accuracy when fixing s and varying k on CIFAR100.  }
\label{fig4}
\end{figure}
  
\begin{figure}
\centering
\includegraphics[scale=0.1]{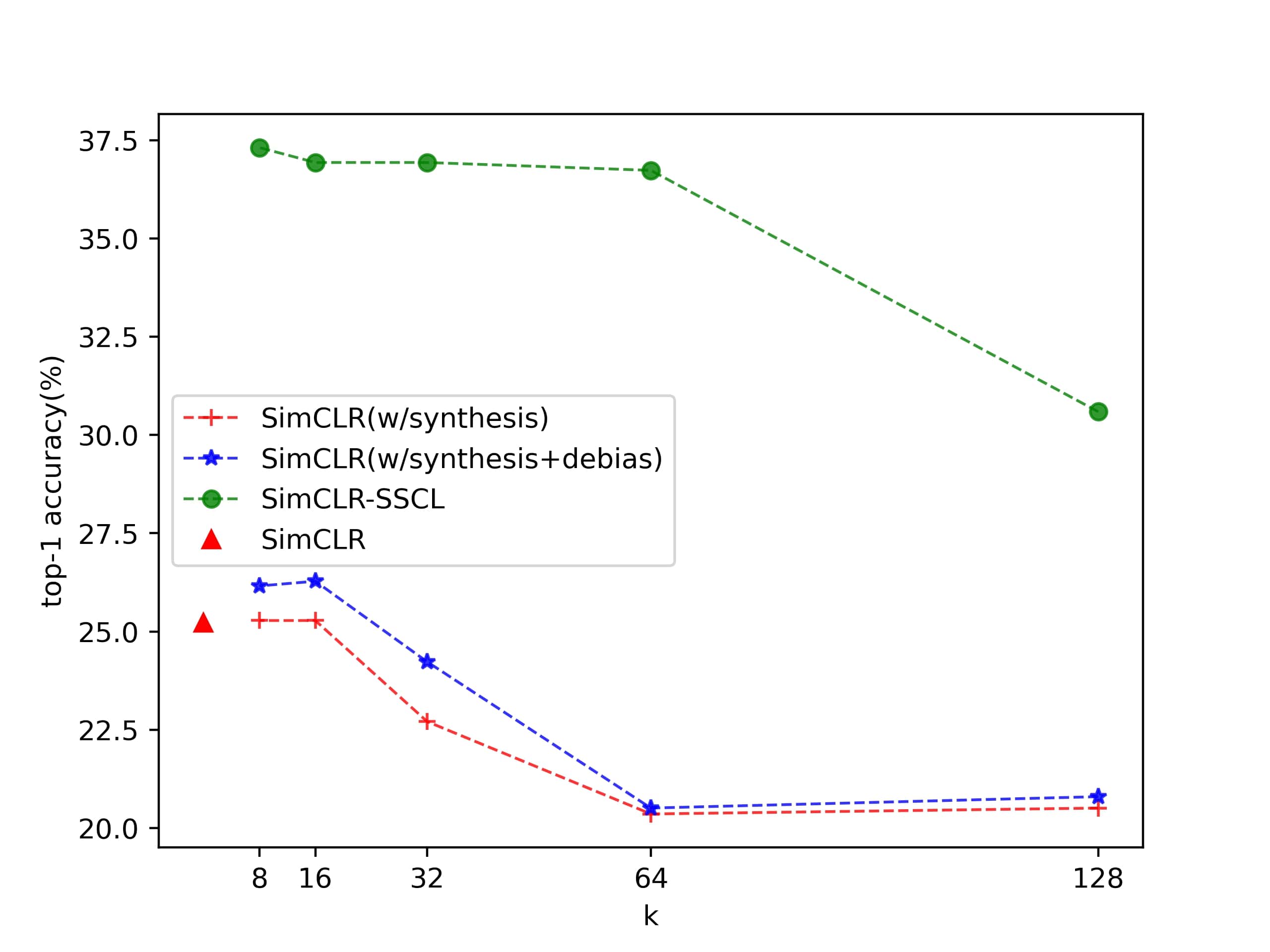}
\vspace{-1em}  
\caption{Top-1 accuracy when fixing s and varying k on TinyImageNet.}
\label{fig5}
\end{figure}

\subsection{Effectiveness of the proposed method}
In order to verify the effectiveness of various aspects of the method we proposed, we carried out experiments on SimCLR \cite{b1} to synthesize hard negative samples and sample negative samples respectively. In Table \ref{table3}, we can observe a notable enhancement in the accuracy of sampling negative samples (w/sampling) . On this basis, we synthesize more hard negative samples and sample them, i.e., SimCLR-SSCL, resulting in further accuracy improvement. However, when we operated only synthesizing hard negative samples on SimCLR (w/synthesis),  the improvement was not obvious and even regressed in CIFAR100.

To explain this phenomenon, we performed experiments by synthesizing different numbers of hard negative samples on CIFAR100 and TinyImageNet. In Figure \ref{fig4} and Figure \ref{fig5}, we fixed s$=$32 and performed ablation experiments to investigate the impact of varying values of k (x-axis). We observed that not the more and harder negative samples are synthesized, the higher the accuracy. When k$>$s, the accuracy of SimCLR (w/synthesis) decreased very significantly. We believe the reasons for this are as follows:

\paragraph{The Effect of False Negative Samples}

There are false negative samples in negative sample set, and directly selecting the most similar may lead to the more severe problem of false negative sample. When k$>$s, if there are false negative samples in the selected negative samples, they may be used several times in subsequent synthesis, resulting in the synthesis of more false negative samples and causing the accuracy to decrease. 

In light of the detrimental effects of false negative samples, we introduced a debiasing approach based on SimCLR (w/synthesis) in our experiments, i.e., SimCLR (w/synthesis+debias), and observed a slight improvement in accuracy. This also shows that the false negative samples in synthetic hard negative samples do affect the results.

\paragraph{The Effect of Randomness}
Due to our experiments' small batch size setting, random selection does not necessarily select negative samples that contain informative content, resulting in certain errors. 

Therefore, it is important to synthesize as few and effective hard negative samples as possible when synthesizing hard negative samples. Our proposed SSCL method further samples the negative samples after synthesizing harder negative samples, eliminates the false negative sample problem through debiasing, and ensures the purity of the negative sample set.
Furthermore, we leverage weights to enhance the contrast between the true negative samples and anchor sample, enabling the true hard negatives to work and providing more semantic information for the network model.
\begin{figure}
\centering
\includegraphics[scale=0.72]{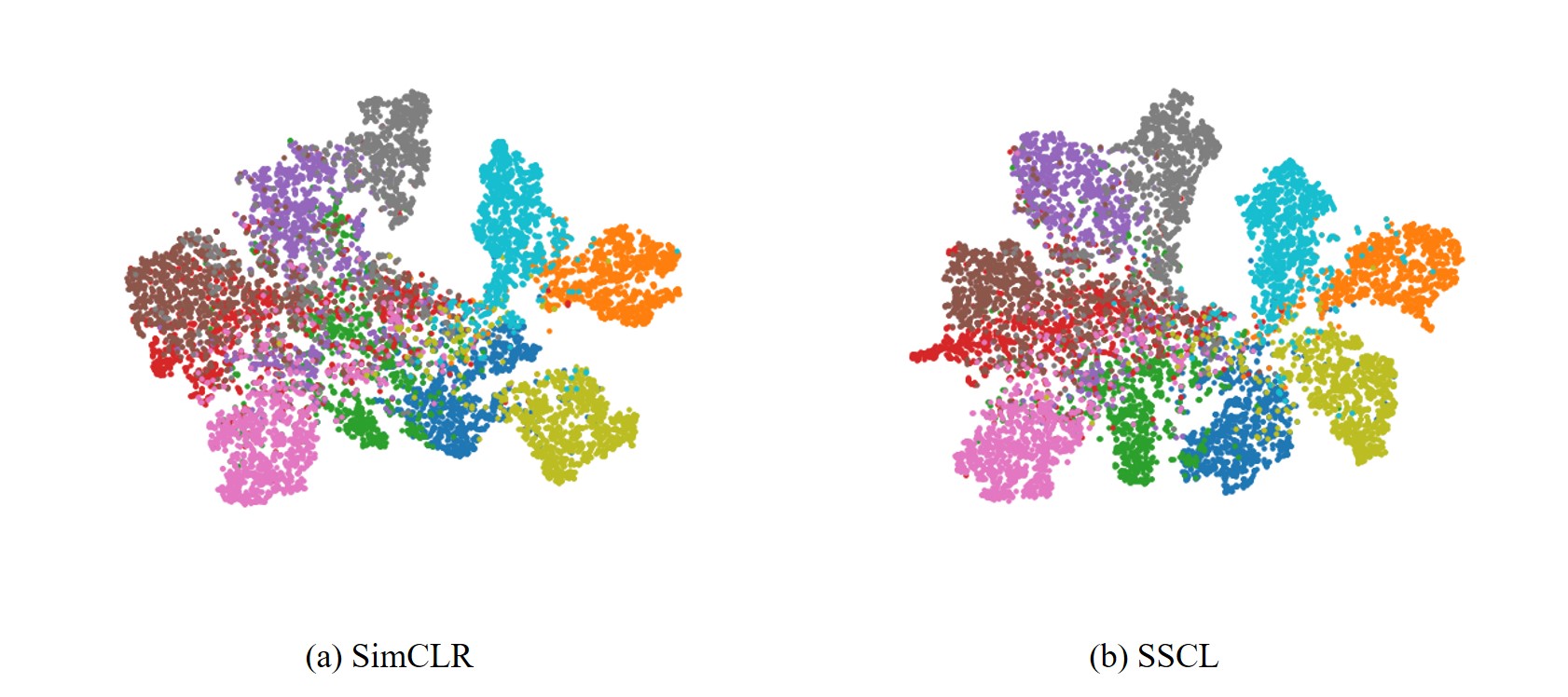}
\caption{t-SNE visualizations feature vectors from CIFAR10 test set. The features are learned by (a) SimCLR and (b) SSCL.  }
\label{fig6}
\end{figure}

Figure \ref{fig6} shows t-SNE \cite{b21} visualizations of the features learned by SimCLR \cite{b1} and SSCL from CIFAR10.
Each color represents a class, SSCL brings about better class separation compared to SimCLR. This shows the instance discrimination ability of SSCL, which solves the problem of false negative sample to some extent.

\begin{table*}[ht]
\caption{Ablation studies for SSCL based on MoCo v2}
\vspace{+1.0em}
\centering
\label{table4}
\begin{tabular}{ccllcllcc}
\hline
\textbf{Method}       & \multicolumn{2}{c}{\textbf{CIFAR10}} &  & \multicolumn{2}{c}{\textbf{CIFAR100}} &  & \multicolumn{2}{c}{\textbf{TinyImageNet}} \\ \cline{2-3} \cline{5-6} \cline{8-9} 
             & top-1        & top-5        &  & top-1         & top-5        &  & top-1           & top-5          \\ \hline
MoCo v2      & 81.45        & 99.19        &  & 47.71         & 78.64        &  & 22.57           & 48.87          \\
w/synthesis  & 82.80        & 99.28        &  & 51.25         & 81.29        &  & 26.32           & 53.22          \\
w/sampling     & 82.90        & 99.33        &  & 51.03         & 81.15        &  & 27.58           & 54.84          \\
MoCo v2-SSCL & \textbf{83.43}        & \textbf{99.33}        &  & \textbf{53.66}         & \textbf{82.43}        &  & \textbf{31.16}           & \textbf{58.14}          \\ \hline
\end{tabular}
\end{table*}

\subsection{Generalization of the proposed method}
Meanwhile, to investigate  the generalization of SSCL, we conducted the same experiments on MoCo v2 \cite{b3}, in which we referred to \cite{b11} for the parameters related to sampling $\beta$=0.2, $\tau$=0. Secondly, we referred to \cite{b8} for the number of negative samples, i.e., s=k=1024. Table \ref{table4} shows that both synthesis and sampling are effective, and SSCL can further enhance accuracy.

Notably, in Table \ref{table4}, MoCov2 (w/synthesis) does not appear as SimCLR (w/synthesis). We assume that MoCo v2 use queue, which can store much more negative samples. Additionally, the selection of negative samples using s=k=1024 may reduce the probability of selecting false negative samples.
The excellent performance of SSCL on MoCo v2 illustrates that our method exhibits good generalizability and can be effectively applied to other frameworks of contrastive learning that require negative samples.

\section{Conclusion}
In this work, a synthetic hard negative samples sampling method is proposed.
By applying synthesis strategy to introduce more and harder negative samples, followed by debiasing them and sampling the true negative samples containing richer semantic information. SSCL exhibits excellent performance on different datasets and can be easily integrated into other existing methods, thereby demonstrating its  effectiveness and generalizability.

Finally, are there more effective criteria for selecting hard negative samples from negative sample set? We also found that the problem of false negative sample still exists. Is there any other way to mitigate the impact of false negative sample while also ensuring the quality of negative samples? These issues raised above are the ones we need to explore further in the future.

\section*{Acknowledgment}

This work was supported by the National Natural Science Foundation of China under Grant No. 61906098.

\bibliography{mybibfile}

\end{document}